\documentclass{article}

\usepackage[normalem]{ulem}
\useunder{\uline}{\ul}{}
\usepackage[table]{xcolor}

\usepackage{amsmath}
\usepackage{amssymb}

\usepackage{booktabs}
\usepackage{amsmath}
\usepackage{amssymb}
\usepackage{makecell}
\usepackage{bbding}
\usepackage{xcolor}
\usepackage{units}
\usepackage{spconf,amsmath,graphicx}
\usepackage{microtype}
\usepackage{subcaption,booktabs}
\usepackage{bm}
\usepackage{enumitem}

\usepackage[pagebackref=false,breaklinks=true,letterpaper=true,colorlinks,bookmarks=false,citecolor=blueish,linkcolor=brickred]{hyperref}

\def\VspaceFa{\vspace{-0.40cm}}
\def\VspaceFb{\vspace{-0.30cm}}

\def\VspaceTb{\vspace{-0.30cm}}

\def\VspacePa{\vspace{-0.30cm}}
\def\VspacePb{\vspace{-0.20cm}}

\def\VspacePc{\vspace{-0.15cm}}
\def\VspacePd{\vspace{-0.10cm}}

\def\VspaceSen{\vspace{0.10cm}}

\definecolor{brickred}{rgb}{0.8, 0.25, 0.33}
\definecolor{blueish}{rgb}{0.0, 0.3, 0.6}

\title{HTNet: Human Topology Aware Network for 3D Human Pose Estimation}

\name{Jialun Cai \quad Hong Liu \quad Runwei Ding\sthanks{Corresponding author: dingrunwei@pku.edu.cn. This work is supported by  National Natural Science Foundation of China (No.62073004), Basic and Applied Basic Research Foundation of Guangdong (No. 2020A1515110370), Shenzhen Fundamental Research Program (GXWD20201231165807007-20200807164903001, No. JCYJ20200109140410340).}
 \quad Wenhao Li \quad Jianbing Wu \quad Miaoju Ban}
\address{Key Laboratory of Machine Perception, Shenzhen Graduate School, Peking University \\
\texttt{\small\{cjl, kimbing.ng, miaoju.ban\}@stu.pku.edu.cn, \{hongliu, dingrunwei, wenhaoli\}@pku.edu.cn}}

\begin{document}
\maketitle

\begin{abstract}
3D human pose estimation errors would propagate along the human body topology and accumulate at the end joints of limbs. Inspired by the backtracking mechanism in automatic control systems, we design an Intra-Part Constraint module that utilizes the parent nodes as the reference to build topological constraints for end joints at the part level.
Further considering the hierarchy of the human topology, joint-level and body-level dependencies are captured via graph convolutional networks and self-attentions, respectively.
Based on these designs, we propose a novel Human Topology aware Network (HTNet), which adopts a channel-split progressive strategy to sequentially learn the structural priors of the human topology from multiple semantic levels: joint, part, and body.
Extensive experiments show that the proposed method improves the estimation accuracy by 18.7\% on the end joints of limbs and achieves state-of-the-art results on Human3.6M and MPI-INF-3DHP datasets. 
Code is available at \href{https://github.com/vefalun/HTNet}{https://github.com/vefalun/HTNet}. 

\end{abstract}
\VspacePd
\vspace{-0.2cm}
\section{Introduction}
\VspacePb
\label{sec:intro}

3D human pose estimation (HPE) from a monocular image is a challenging task, which has been widely applied in the sub-tasks of human analysis, such as action recognition~\cite{icassp_action} and person re-identification~\cite{wangtao}. 
Benefiting from the effective 2D HPE framework~\cite{chen2018cascaded}, most recent works focus on the 2D-to-3D lifting pipeline~\cite{simplebaseline}, which firstly estimates 2D poses from a single image and then lifts them to 3D keypoints. 
Unlike image-based tasks, the 2D-to-3D pose lifting task takes inherently sparse and structural 2D joint coordinates as inputs. Therefore, it is critical to take full advantage of the structural priors of the human topology. 

\begin{figure}[t]
    \centering
    \includegraphics[width=\linewidth]{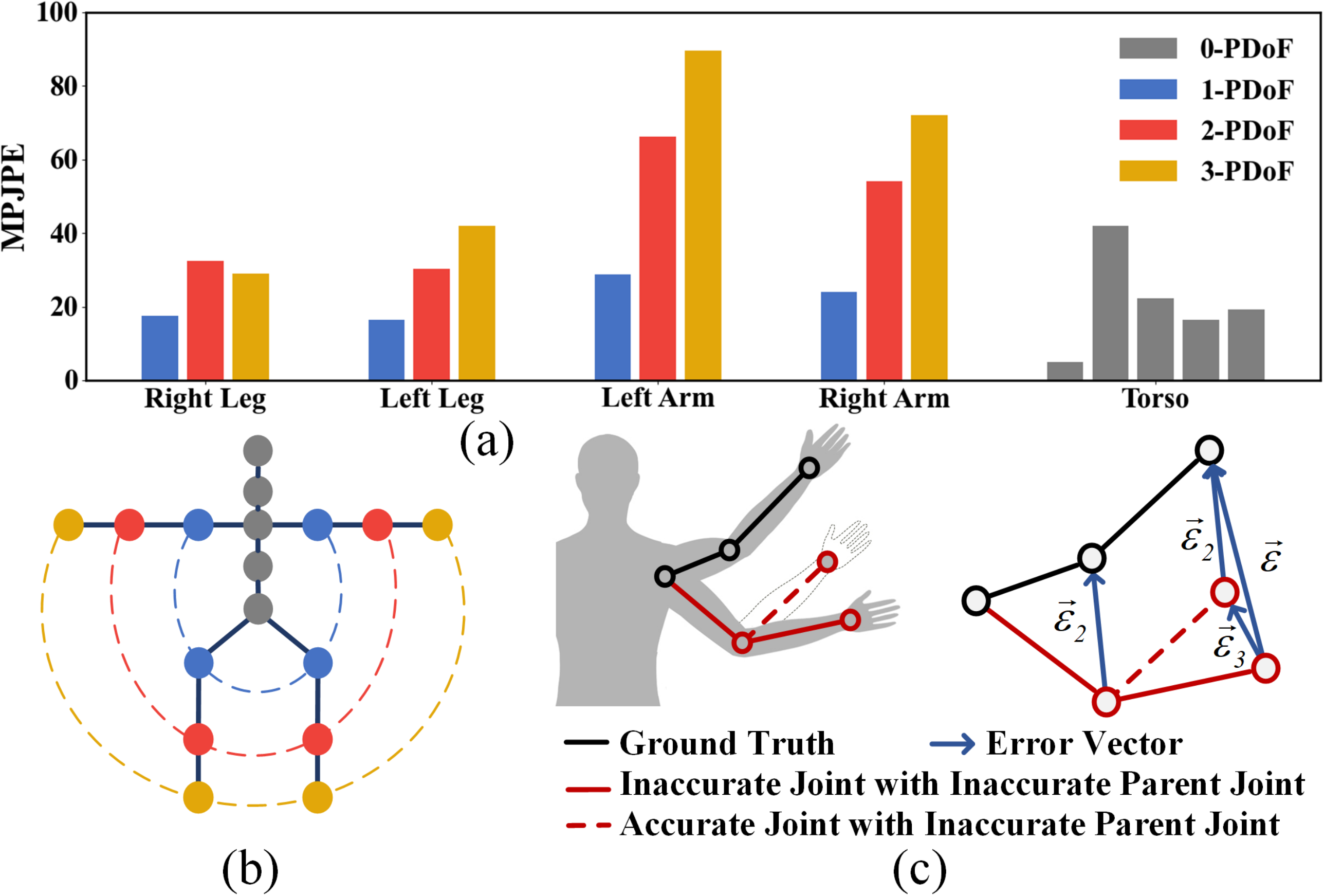}
    \VspaceFa
    \vspace{-0.30cm}
    \caption
    {
        (a) Distribution of estimation errors; (b) Joints with different PDoFs;  (c) Physiological explanation of error accumulation at the end joints (\textit{e.g.}, wrist): $\vec{\bm{\varepsilon}} = \bm{\vec{\varepsilon}}_\textit{3} + \bm{\vec{\varepsilon}}_\textit{2}$.
    }
    \VspaceFb
    \label{fig:motivation}
\end{figure}

Recently, many works~\cite{stgcn,mgcn,li2022graphmlp} have focused on the most related local topology by modeling the correlations among body joints via Graph Convolutional Networks (GCNs), while others~\cite{lightpose,poseformer,mhformer} have focused on the less related global contexts via the Multi-head Self-Attention (MSA) of Transformer. 
However, the skeleton representations of the human body cannot be sufficiently captured by these methods due to the complex topological relationships among joints, leading to dramatically wrong estimation, especially at the end joints of limbs.
As shown in Fig.~\ref{fig:motivation} (a), the previous state-of-the-art model~\cite{mgcn} suffers from significant estimation errors on joints with high PDoFs, and here we define the Part Degree of Freedom (PDoF) of joints via the distance to the torso (see Fig.~\ref{fig:motivation} (b)). To address these issues, we explore the human topology from the following two aspects: 

\textit{\textbf{(i)}} \textbf{\textit{Error accumulation}}: 
Since the human body is a linkage structure that limb joints highly depend on their parent nodes~\cite{wu2021limb} shown in Fig.~\ref{fig:motivation} (c), the estimation errors would accumulate from the central hip in the torso (root joint) to the end joints of limbs. 
In automatic control systems, backtracking to the previous points with fewer errors and leveraging their kinematics priors can effectively alleviate the problem of error accumulation~\cite{ backtracking,backtracking_1}.
Inspired by it, an Intra-Part Constraint (IPC) is designed to take intra-part parent joints as the reference to constrain the joints with higher PDoFs. 

\begin{figure*}[t]
\centering
\VspaceFa
\includegraphics[width=0.92\textwidth]{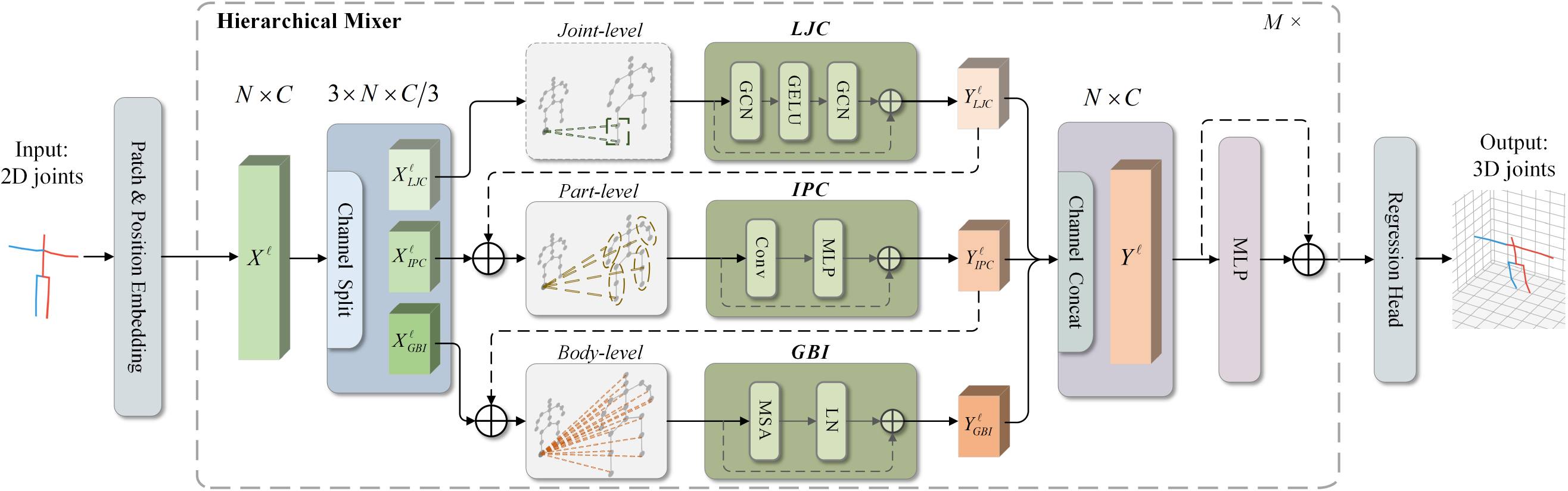}
\caption
{
Overview of the proposed HTNet. The HTNet consists of $M$ stacked Hierarchical Mixers. Each mixer consists of three modules (LJC, IPC, and GBI), which can extract features at joint, part, and body levels.
}
\vspace{-0.5cm}
\label{fig:fig2}
\end{figure*}

\textit{\textbf{(ii)}} \textbf{\textit{Hierarchical structure}}:
The Joint motion is closely linked to the hierarchical organization of human topology, which includes joint~\cite{stgcn}, part~\cite{kundu2020self,xue2022boosting}, and body~\cite{poseformer} levels. Driven by this natural hierarchy, we present a novel Human Topology aware Network (HTNet) that learns human topology representations at multiple levels (see Fig.~\ref{fig:fig2}). 
Specifically, HTNet consists of alternating hierarchical mixers, each composed of three modules.
At the joint level, we design a Local Joint-level Connection (LJC) based on GCN to model the physical connections between adjacent joints;
At the part level, the IPC in (i) provides the constraints for intra-part joints, such that they have similar athletic trends;
At the body level, the Global Body-level Interaction (GBI) based on MSA extracts global features among inter-part joints. 
Notably, the hierarchical mixer adopts a channel-split progressive design, which achieves a win-win scenario by providing more expressive features from various levels while keeping a small model size. 

In summary, the main contributions are as follows: 
\begin{itemize}[itemsep=0pt,topsep=0pt,parsep=0pt]
\item We propose a Human Topology aware Network (HTNet) with a channel-split progressive design to learn human topology dependencies at joint, part, and body levels.
\item To alleviate the error accumulation, we design an Intra-Part Constraint (IPC) module that utilizes topological constraints of parent nodes to reduce errors of end joints.

\item Extensive experiments demonstrate the effectiveness of HTNet, which achieves state-of-the-art results on both Human3.6M and MPI-INF-3DHP benchmark datasets. 
\end{itemize}

\VspacePa
\section{Methodology}
\VspacePb
\label{sec:method}
\VspacePc
\subsection{Overview}
\VspacePd
\label{sec:2.1}
The overview of HTNet is depicted in Fig.~\ref{fig:fig2}. Given the 2D coordinates $X {\in} \mathbb{R}^{N\times 2}$, we firstly map $X$ to a high dimension 
${X}' {\in} \mathbb{R}^{N\times C}$ via the patch embedding, where $N$ is the number of joints and $C$ is the channel dimensions. Then, we embed ${X}'$ with a learnable positional matrix $E_{pos} {\in} \mathbb{R}^{N\times C}$ to obtain the embedded features $X^\ell$, where $\ell$ is the index of hierarchical mixers. Next, the $X^\ell$ is split into three parts: $X^\ell_{\mbox{\textit{\tiny LJC}}}$, $X^\ell_{\mbox{\textit{\tiny IPC}}}$, and $X^\ell_{\mbox{\textit{\tiny GBI}}}$, which have equal dimensions $C'{=}C/3$ and are fed into corresponding modules: LJC, IPC, GBI.
These modules construct the human topology from multiple semantic levels: joint, part, and body.
Finally, we aggregate hierarchical representations and use a linear layer as the regression head to produce the 3D coordinates $Y {\in} \mathbb{R}^{N\times 3}$. 
Each module and the structure of HTNet will be introduced in the following sections.

\VspacePc
\subsection{Local Joint-level Connection}
\VspacePd
\label{sec:2.2}
Local Joint-level Connection (LJC) is a GCN-based architecture~\cite{mgcn}. The graph-structured human skeleton can be defined as $G {=} (V,A)$, where $V$ is a set of $N$ joints and $A {\in} \left \{ 0,1 \right \}^{N\times N }$ is an adjacency matrix representing the connection relations between joints. 
Given the input $X^{\ell}_{\mbox{\textit{\tiny LJC}}} {\in} \mathbb{R}^{N\times C'}$, features of neighboring joints can be aggregated to $Y^{\ell}_{\mbox{\textit{\tiny LJC}}}$ by GCN:
\begin{equation}
\setlength{\abovedisplayskip}{5pt}
    \begin{array}{l}
\mbox{\textit{GCN}}(X^\ell_{\mbox{\textit{\tiny LJC}}}) = \tilde{D}^{-\frac{1}{2}}\tilde{A} \tilde{D}^{-\frac{1}{2}}X^{\ell}_{\mbox{\textit{\tiny LJC}}}W, \\
Y^{\ell}_{\mbox{\textit{\tiny LJC}}} = X^{\ell}_{\mbox{\textit{\tiny LJC}}} + \mbox{\textit{GCN}}(\sigma(\mbox{\textit{GCN}}(X^{\ell}_{\mbox{\textit{\tiny LJC}}}))) ,
\label{eq1}
    \end{array}
\setlength{\belowdisplayskip}{2pt}
\end{equation}
where $\tilde{A}{=}A{+}I$, $\tilde{D}$ is the diagonal node degree matrix, $W$ is the weight matrix, $\sigma$ denotes the $\mbox{\textit{GELU}}$ activation function.

\VspacePc
\VspacePc
\subsection{Intra-Part Constraint}
\begin{figure}[t]
    \centering
    \includegraphics[width=0.95\linewidth]{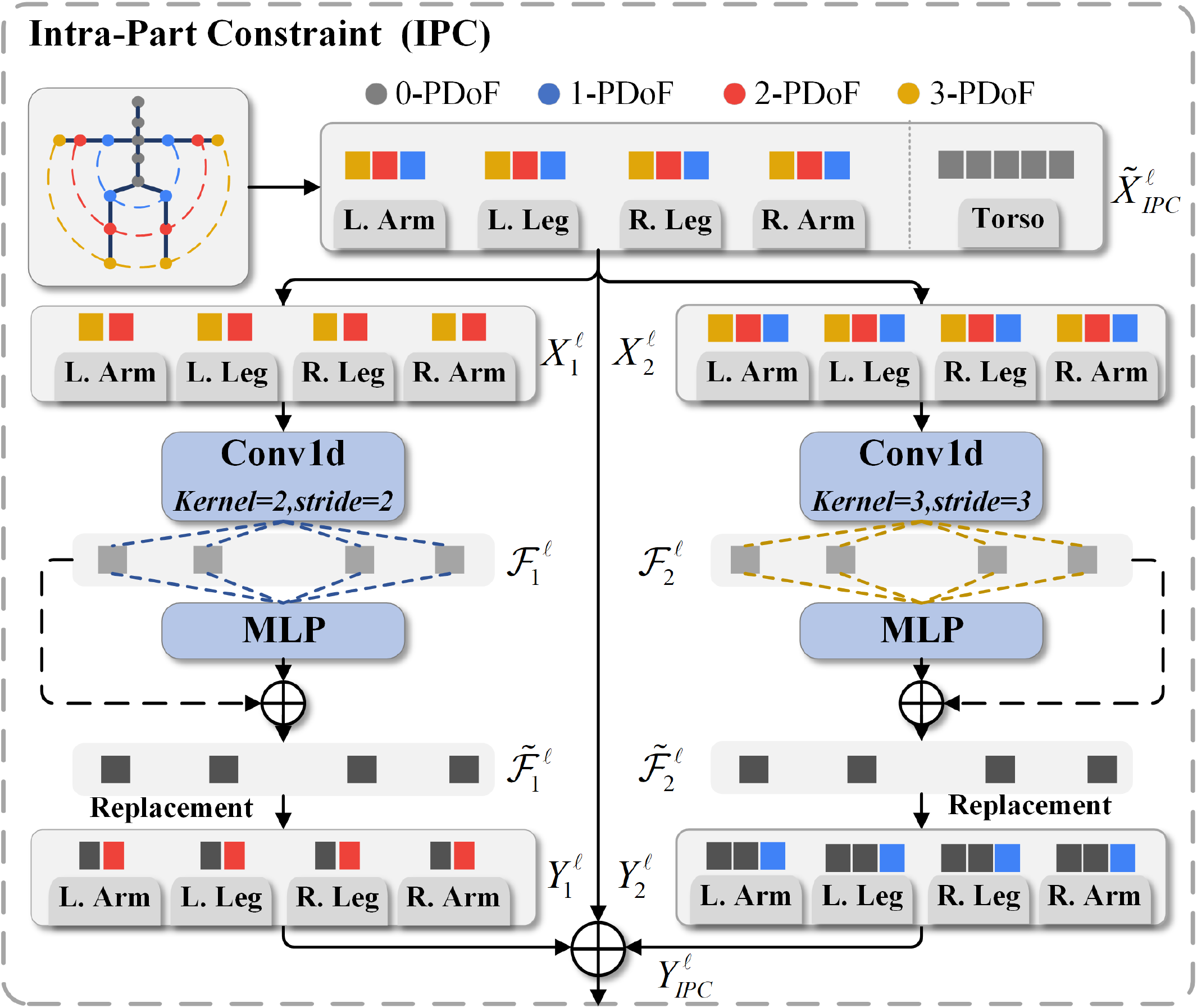}
    \vspace{-0.1cm}
    \caption
    {
        Structure of Intra-Part Constraint (IPC) module.
    }
    \VspaceFb
    \vspace{-0.1cm}
    \label{fig:fig3}
\end{figure}

\VspacePd
\label{sec:IPC}
To alleviate the error accumulation, we design an Intra-Part Constraint (IPC) module (see Fig.~\ref{fig:fig3}), which aims to conduct the constraint among intra-part joints with different PDoFs. 

Given the input $\tilde{X}^\ell_{\mbox{\textit{\tiny IPC}}} = X^\ell_{\mbox{\textit{\tiny IPC}}} + Y^\ell_{\mbox{\textit{\tiny LJC}}}$, topological constraints performs in two sets of joints: \textit{(\romannumeral1)} $X^\ell_{\textit{1}} {\in} {\mathbb{R}^{8 \times C'}}$ consists of 2-PDoF and 3-PDoF joints; \textit{(\romannumeral2)} $X^\ell_{\textit{2}} {\in} {\mathbb{R}^{12 \times C'}}$ consists of 1-PDoF, 2-PDoF, and 3-PDoF joints. Next, $X^\ell_{\textit{j}},j {\in} \{1,2\}$ are fed into two convolution layers to generate limb features $\mathcal{F}^\ell_{j}$. Then, a channel MLP~\cite{mlpmixer} is adopted to aggregate information between different channels for each limb feature:
\begin{equation}
\setlength{\abovedisplayskip}{5pt}
    \begin{array}{l}
    \mathcal{F}^\ell_{j}=\sigma(\mbox{\textit{Conv}}_{j}(X^\ell_{j})), j\in\{1,2\},\\[0.1cm] 
    \mathcal{\tilde{F}}^\ell_{j}=\mathcal{F}^\ell_{j} + \mbox{\textit{MLP}}(\mbox{\textit{LN}}(\mathcal{F}^\ell_{j})),
    \end{array}
\setlength{\belowdisplayskip}{2pt}
\end{equation}
where $\mathcal{\tilde{F}}^\ell_{j}$ are aggregated limb features, $LN(\cdot)$ denotes Layer Normalization, $\mbox{\textit{Conv}}_{\textit{1}}$ and $\mbox{\textit{Conv}}_{\textit{2}}$ are convolution layers with kernel sizes of 2 and 3, respectively. 
Based on the above $\mathcal{\tilde{F}}^\ell_{\textit{j}}$, we construct two topological constraints $\mathcal{R}_{j}$ in $X^\ell_\textit{j}$: 
\textit{(\romannumeral1)} In $\mathcal{R}_{1}$, 3-PDoF joints in $X^\ell_\textit{1}$ are replaced with $\mathcal{\tilde{F}}^\ell_{\textit{1}}$;
\textit{(\romannumeral2)} In $\mathcal{R}_{2}$, 3-PDoF and 2-PDoF joints in $X^\ell_\textit{2}$ are replaced with $\mathcal{\tilde{F}}^\ell_{\textit{2}}$.
Then, the unprocessed joints are filled into the replaced $X^\ell_j$, and $Y^\ell_{\mbox{\textit{\tiny IPC}}}$, the final output of IPC, can be represented as follows:
\begin{equation}
\setlength{\abovedisplayskip}{5pt}
    Y^\ell_{\mbox{\textit{\tiny IPC}}} = \tilde{X}^\ell_{\mbox{\textit{\tiny IPC}}} + \mathcal{R}_{1}(\tilde{X}^\ell_{\mbox{\textit{\tiny IPC}}}, \mathcal{\tilde{F}}^\ell_{1}) + \mathcal{R}_{2}(\tilde{X}^\ell_{\mbox{\textit{\tiny IPC}}}, \mathcal{\tilde{F}}^\ell_{2}).
\setlength{\belowdisplayskip}{4pt}
\end{equation}

Since the limb features $\mathcal{\tilde{F}}^\ell_{j}$ are the motion representation of limbs, which contains both high-PDoF joints and their parent joints. Therefore, replacing the high-PDoF joints with $\mathcal{\tilde{F}}^\ell_{j}$ can form topological constraints and produce reasonable 3D pose estimations to mitigate the error accumulation.

\VspacePc
\VspacePc
\VspacePc

\subsection{Global Body-level Interaction}
\VspacePd
\vspace{-0.05cm}
\label{sec:2.4}
Global Body-level Interaction (GBI) module is built by the self-attention of Transformer, which can capture global contexts across the whole human body~\cite{mhformer,GraFormer}.
Given $\tilde{X}^\ell_{\mbox{\textit{\tiny GBI}}} = X^\ell_{\mbox{\textit{\tiny GBI}}} + Y^\ell_{\mbox{\textit{\tiny IPC}}}$ as input, the GBI module can be formulated as:
\begin{equation}
\setlength{\abovedisplayskip}{3pt}
    \begin{array}{l}
    H_i=\mbox{\textit{Softmax}}({Q^{\ell}{K^{\ell}}^{T}}{/}{\sqrt{C'}})V^{\ell},i\in\{1,...,h\}\\[0.05cm]
        \mbox{\textit{MSA}}(\tilde{X}^\ell_{\mbox{\textit{\tiny GBI}}})=\mbox{Concat} (H_1,H_2,...,H_h)W_{out}, \\[0.05cm]
        \tilde{Y}^\ell_{\mbox{\textit{\tiny GBI}}} = \tilde{X}^\ell_{\mbox{\textit{\tiny GBI}}} + \mbox{\textit{LN}}(\mbox{\textit{MSA}}(\tilde{X}^\ell_{\mbox{\textit{\tiny GBI}}})),
    \end{array}
\setlength{\belowdisplayskip}{1pt}
\label{eq2}
\end{equation}
where $h$ is the number of attention heads, $Q^{\ell},K^{\ell},V^{\ell}$ are query, key, and value matrices, which are calculated from $\tilde{X}^\ell_{\mbox{\textit{\tiny GBI}}}$ by linear transformations, $\tilde{Y}^\ell_{\mbox{\textit{\tiny GBI}}}$ is the output of the GBI module.

\VspacePc
\VspacePc
\VspacePc

\subsection{Network Structure}
\VspacePd
\vspace{-0.05cm}
\label{sec:2.5}
Although hierarchical representations of human topology can be learned with LJC, IPC and GBI, it is challenging to design a connection structure for such three modules: the series structure (\textit{i.e.}, three modules are connected sequentially) will increase the model size, and the parallel structure (\textit{i.e.}, three modules divide the channels equally and work in parallel) lacks feature interactions among various levels. 

Inspired by Res2Net~\cite{res2net}, we design the hierarchical mixer with a channel-split progressive structure.
Specifically, channels are split into three parts and fed into LJC, IPC, and GBI.
Meanwhile, the output of the previous module is added to the input of the next module as residual-like connections for learning the human topology from local to global.
Then, outputs of the above three modules are concatenated to generate $Y^\ell {\in} \mathbb{R}^{N\times C}$, and $Y^\ell$ is fed to a channel MLP block:
\begin{equation}
\setlength{\abovedisplayskip}{3pt}
\begin{array}{l}
    Y^\ell = \mbox{Concat}(Y^\ell_{\mbox{\textit{\tiny LJC}}},Y^\ell_{\mbox{\textit{\tiny IPC}}},Y^\ell_{\mbox{\textit{\tiny GBI}}}),\\
    \tilde{Y}^\ell = Y^\ell + \mbox{\textit{MLP}}(LN(Y^\ell)).
    \label{eq4}
\end{array}
\setlength{\belowdisplayskip}{1pt}
\end{equation}

Such a process can aggregate features from three different levels and obtain a more expressive representation.

\VspacePa
\section{Experiments}
\label{sec:Experiments}
\subsection{Datasets and Evaluation Metrics}
\VspacePd
\noindent \textit{\textbf{Human3.6M}}~\cite{ionescu2013human3} is the largest indoor dataset for 3D HPE. It has 3.6 million images and 11 professional actors. Following~\cite{mgcn,GraFormer}, Mean Per Joint Position Error (MPJPE) and Procrustes MPJPE (P-MPJPE) are adopted as metrics.
\VspaceSen

\noindent \textit{\textbf{MPI-INF-3DHP}}~\cite{mehta2017monocular} consists of 1.3 million images from indoor and outdoor scenes. There are three different scenes in its test set: studio with a green screen (GS), studio without a green screen (noGS), and outdoor scene (Outdoor). Following~\cite{mgcn,GraFormer,liu2020comprehensive}, MPJPE, Percentage of Correct Keypoint (PCK), and Area Under Curve (AUC) are adopted as metrics.

\VspacePc
\subsection{Implementation Details} 
\VspacePd
We implement our method using Pytorch and train the model for 30 epochs with batch size 512. 
The proposed HTNet consists of 3 stacked hierarchical mixers.
Note that the channel dimension of inputs should be divisible by 24 (\textit{e.g.}, 240, 360) because there are eight heads in the MSA and three split channels in hierarchical mixers. 
The $L_{2}$ loss is utilized to minimize the errors between predictions and ground truth.

\begin{table*}[t]
    \begin{subtable}[t]{0.32\textwidth}
            \footnotesize
            \centering
            \setlength{\tabcolsep}{0.70mm} 
          \begin{tabular}{@{}l|ccc}
          \toprule
            Method &\textbf{P1 (CPN)}  &\textbf{P2 (CPN)}  &\textbf{P1 (GT)}\\
          \midrule
          Martinez \textit{et al.}~\cite{simplebaseline}   &62.9 &47.7 &45.5 \\
          Ci \textit{et al.}~\cite{ci2019optimizing}  &52.7 &42.2 &36.3 \\ 
          Liu \textit{et al.}~\cite{liu2020comprehensive}   &52.4 &41.2 &37.8 \\
          Xu \textit{et al.}~\cite{xu2021graph}   &51.9 &- &35.8 \\
          Zhao \textit{et al.}~\cite{GraFormer}  &51.8 &- &{35.2} \\ 
          Cai \textit{et al.}~\cite{stgcn}$(\dagger)$   & 50.6 & 40.2& 38.1\\
          Zeng \textit{et al.}~\cite{zeng2020srnet}  &49.9 &39.4 &36.4\\
          Zou \textit{et al.}~\cite{mgcn}$(\dagger)$ &{49.4} &{39.1} &37.4 \\
          \midrule
          \rowcolor[HTML]{DADADA}
          {HTNet (Ours) }  &{48.9} &{39.0} &{34.0} \\
          \rowcolor[HTML]{DADADA}
          {HTNet (Ours) $(\dagger)$}  &\textbf{47.6} &\textbf{38.6} &\textbf{31.9} \\
          \bottomrule
          \end{tabular}
          \caption{Quantitative comparison on Human3.6M.}
          \label{table:human}
          \VspaceTb
    \end{subtable}
    \hfill
    \enspace
    \begin{subtable}[t]{0.35\textwidth}
        	\centering
            \footnotesize
            \setlength{\tabcolsep}{0.50mm} 
        	\begin{tabular}{l|ccccc}
        	\toprule
        	Method  &\textbf{ GS}$\uparrow$ & \textbf{noGS}$\uparrow$ & \textbf{Outdoor}$\uparrow$ & \textbf{PCK}$\uparrow$ & \textbf{AUC}$\uparrow$        \\
        	\midrule
        	Martinez \textit{et al.}~\cite{simplebaseline}  & 49.8 &42.5 &31.2 &42.5 & 17.0 \\
        	Mehta \textit{et al.}~\cite{mehta2017monocular}  & 70.8 &62.3 &58.5 &64.7 &31.7 \\
        	Ci \textit{et al.}~\cite{ci2019optimizing}  & 74.8 &70.8 &77.3 &74.0 &36.7 \\
        	Zhou \textit{et al.}~\cite{zhou2017towards}  & 75.6 &71.3 &80.3 &75.3 &38.0 \\
            Zeng \textit{et al.}~\cite{zeng2020srnet}   &- &-  &80.3 &77.6 &43.8  \\
            Liu \textit{et al.}~\cite{liu2020attention}  &77.6 &80.5 &80.1 &79.3 &45.8 \\
            Zeng \textit{et al.}~\cite{zeng2021learning}  &- &-  &84.6 &82.1 &46.2  \\
            Xu \textit{et al.}~\cite{xu2021graph}  &81.5 &81.7 &75.2 &80.1 &45.8 \\
            Zou \textit{et al.}~\cite{mgcn} &{86.4} &{86.0} &{85.7} &{86.1} &{53.7} \\
            \midrule
            \rowcolor[HTML]{DADADA}
        	HTNet (Ours) &\textbf{86.9} &\textbf{86.2} &\textbf{85.9} &\textbf{86.7}  &\textbf{54.1}	\\
        	\bottomrule
        	\end{tabular}
            \caption{Quantitative comparison on MPI-INF-3DHP.}
        	\label{tab:3dhp}
        	\VspaceTb
     \end{subtable}
    \hfill
    \enspace
    \begin{subtable}[t]{0.30\textwidth}
            \footnotesize
            \centering
            \setlength{\tabcolsep}{0.35mm}
            \begin{tabular}{@{}l|cc|c@{}}
            \toprule
            Method & \textbf{Frame}  & \textbf{Param }& \textbf{MPJPE}  \\
            \midrule
            Pavllo \textit{et al.}~\cite{videopose} &{9}  &{4.4M} &{49.8} \\
            
            Cai \textit{et al.}~\cite{stgcn}  &{7}  &{5.1M} &{48.8} \\
            Zheng \textit{et al.}~\cite{poseformer}  &{9}  &{9.6M} &{49.9} \\
            
            Li \textit{et al.}~\cite{mhformer}  &{9}  &{18.9M} &{47.8} \\
            Chen \textit{et al.}\cite{chen2021anatomy}  &{9} &{18.2M} &{46.3} \\

            \rowcolor[HTML]{DADADA}
            HTNet-S (Ours) &{9}  &\textbf{3.0M} &47.2 \\
            \midrule
            Pavllo \textit{et al.}~\cite{videopose}  &{27}  &{8.6M} &{48.8} \\
            Liu \textit{et al.}~\cite{liu2020attention}  &{27} &{5.7M} &{48.5} \\
            Zheng \textit{et al.}~\cite{poseformer}  &{27}  &{9.6M} &{47.0} \\
            
            \rowcolor[HTML]{DADADA}
            HTNet-L (Ours) &{27}  &10.1M &\textbf{46.1} \\
            \bottomrule
            \end{tabular}
            \caption{Comparison with temporal methods.} 
            \label{table:frame}
            \VspaceTb
     \end{subtable}    
     \caption{\textbf{(a)} Quantitative comparison on Human3.6M under Protocol \#1 (MPJPE) and Protocol \#2 (P-MPJPE). The 2D keypoints detected by CPN and the ground truth of 2D poses are used as inputs. $(\dagger)$ - adopts the same refinement module as~\cite{stgcn,mgcn}. \textbf{(b)} Quantitative comparison on MPI-INF-3DHP. \textbf{(c)} Quantitative comparison with temporal methods on Human3.6M.}
     \label{tab:temps}
     \VspaceTb
\end{table*}

\VspacePc
\subsection{Method Comparison}
\VspacePd
\label{comparison}
\noindent \textbf{Comparison on Human3.6M.} 
We compare HTNet with state-of-the-art methods on Human3.6M. 
As shown in Tab.~\ref{tab:temps} (a), the 2D keypoints from CPN~\cite{chen2018cascaded} are used as inputs, and our model achieves the best results under both MPJPE (47.6\textit{mm}) and P-MPJPE (38.6\textit{mm}).
Besides, using the ground truth 2D joints as inputs, HTNet also performs best, outperforming GraFormer~\cite{GraFormer} by $9.3\%$ (31.9\textit{mm} vs. 35.2\textit{mm}).
\VspaceSen

\noindent \textbf{Comparison on MPI-INF-3DHP.} 
To evaluate the generalization capability of our approach, we train HTNet on the Human3.6M and test it on the MPI-INF-3DHP directly. As shown in Tab.~\ref{tab:temps} (b), HTNet achieves the best performance on all scenes, which verifies the strong generalization of our method in unseen environments. 
\VspaceSen

\noindent \textbf{Comparison with Temporal Methods.}
To explore the portability of HTNet in video-based 3D HPE, we compare with some methods~\cite{stgcn,poseformer,mhformer,liu2020attention,videopose,chen2021anatomy} with similar frames as inputs, and Tab.~\ref{tab:temps} (c) shows that HTNet-S ($C {=} 240$) can achieve competitive performance under MPJPE (47.2\textit{mm}) with much fewer parameters. 
Notably, we find that the channel dimension will be the bottleneck to the performance gains. 
Thus, HTNet-L is built with larger dimensions ($C {=}720$) and performs the best (46.1\textit{mm}). The above experiments prove that our HTNet can work well with video sequence inputs.

\begin{figure}
\begin{minipage}[b]{.48\linewidth}
    \footnotesize
    \centering
    \setlength{\tabcolsep}{0.30mm} 
	\begin{tabular}{lccc|cc}
		\toprule
		& LJC & \makecell IPC & \makecell[c]{GBI}  & Param   & MPJPE\\
		\midrule
		Baseline &   &   &\Checkmark                  	& 2.2M  & 52.2\\
		&\Checkmark  &   &                              & 2.6M & 50.4\\
		& \Checkmark & & \Checkmark                      & 2.5M  & 50.2\\
		&        & \Checkmark  & \Checkmark             & 3.4M  & 51.8\\
		& \Checkmark       & \Checkmark   &              & 3.8M  & 49.3\\
		\midrule
        Parallel &\Checkmark    &\Checkmark  & \Checkmark   &3.0M  &50.0\\
		Serial &\Checkmark    &\Checkmark  & \Checkmark    &7.2M  &49.2\\
		\midrule
		\rowcolor[HTML]{DADADA}
		HTNet   & \Checkmark& \Checkmark& \Checkmark  & 3.0M  & \textbf{48.9}\\
		\bottomrule  
	\end{tabular}%
    	\captionof{table}{Ablation study for components and structures.}
    \VspaceTb
	\label{tab:ablation_component}
\end{minipage}
\enspace
\begin{minipage}[b]{.48\linewidth}
    \centering
    \includegraphics[width=\linewidth]{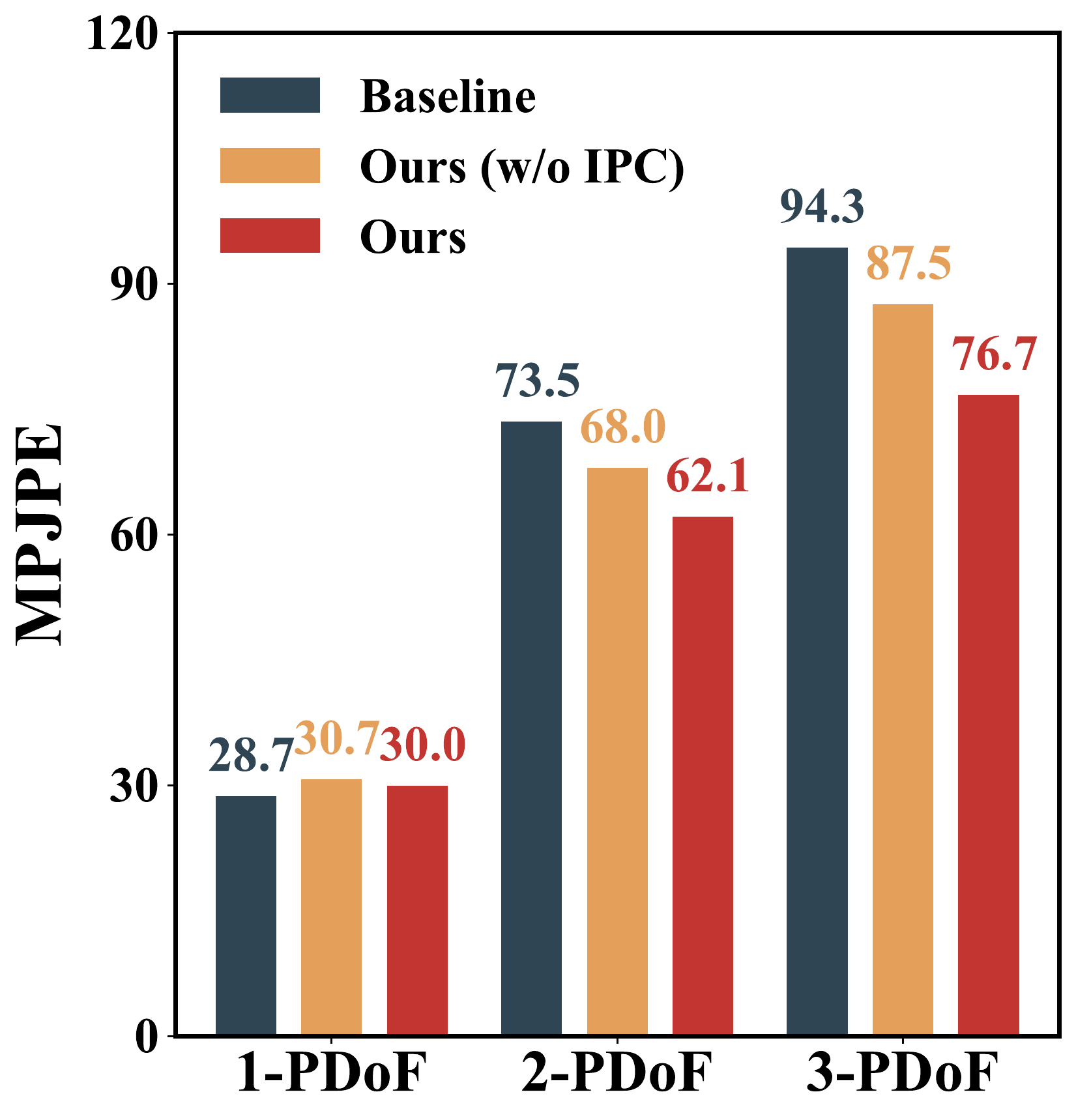}
    \VspaceFa
    \caption{Ablation study for the effect of IPC.}
    \VspaceFb
    \label{fig_IPC}
\end{minipage}%
\VspaceFb
\end{figure}

\VspacePc
\subsection{Ablation Study}
\VspacePd
Tab.~\ref{tab:ablation_component} shows the results of our method under different components and structures on the Human3.6M dataset. For a fair comparison, we maintain the consistency of the parameters by keeping the same dimension channels ($C {=} 240$).

\noindent \textbf{Impact of Model Components.}
As shown in the top part of Tab.~\ref{tab:ablation_component}, HTNet with only a single-level module cannot achieve satisfactory performance, while the combination of all proposed modules performs the best via learning representations from various levels. 
Notably, IPC cannot work individually due to no operations on 0-PDoF and 1-PDoF joints. To further investigate the influence of IPC, we categorize limb joints by PDoFs and calculate the average MPJPE of joints in each category in Fig.~\ref{fig_IPC}.
Compared with the baseline (the HTNet with only GBI), the average MPJPE of all joints decreased by 3.8\% (50.2\textit{mm} vs. 52.2\textit{mm}) via introducing LJC. By further integrating IPC into HTNet, the MPJPE would decreased by 6.3\% (48.9\textit{mm} vs. 52.2\textit{mm}), while MPJPE on 2-PDoF and 3-PDoF joints noticeably reduces by $15.5\%$ and $18.7\%$, respectively.
Such a significant reduction in the estimation errors of end joints can be attributed to the number of topological constraints they obtain shown in Sec.~\ref{sec:IPC}: \textit{(i)} 3-PDoF joints are constrained by both 1-PDoF and 2-PDoF joints; \textit{(ii)} 2-PDoF joints are only constrained by 1-PDoF joints; \textit{(iii)} 1-PDoF joints are not subject to any constraints. 
\vspace{0.05cm}

\noindent \textbf{Impact of Model Structures.}
As for the structures (middle part of Tab.~\ref{tab:ablation_component}), the serial structure exhibits competitive performance (49.2\textit{mm}) due to the local-to-global learning; the parallel structure can reduce parameters from 7.2M to 3.0M due to the channel-split structure. The channel-split progressive structure of HTNet adopts residual connections and combines the advantages of these two designs, which performs the best (48.9\textit{mm}) and maintains a small model size (3.0M).

\begin{figure}[t]
    \centering
    \includegraphics[width=0.95\linewidth]{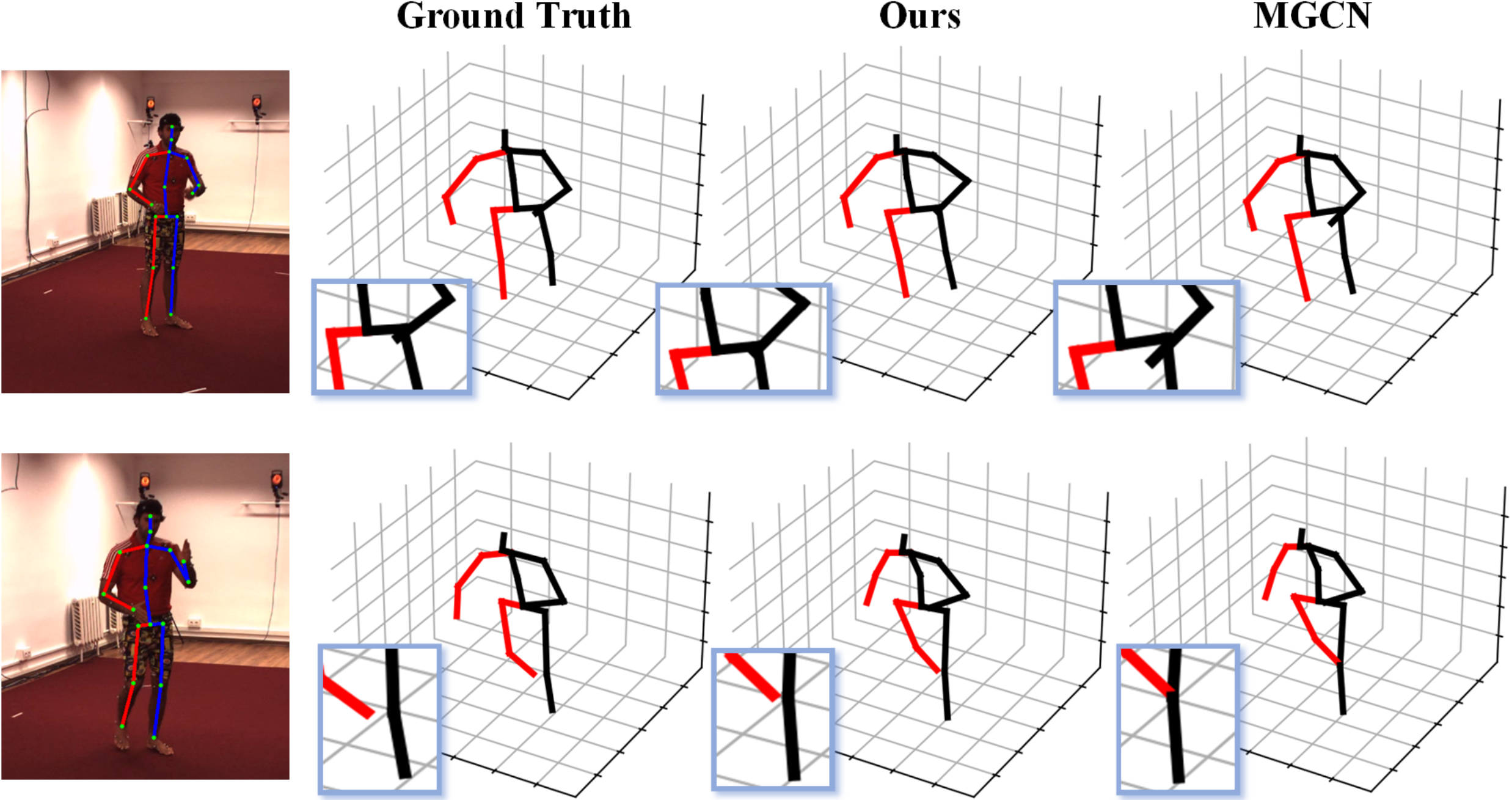}
    \caption
    {
        Qualitative comparison on Human3.6M.
    }
    \label{fig_qua}
    \vspace{-0.1cm}
\end{figure}

\VspacePc
\subsection{Visualization}
\VspacePd
Fig.~\ref{fig_qua} provides qualitative comparisons on Human3.6M between HTNet and the state-of-the-art approach, \textit{i.e.}, MGCN~\cite{mgcn}.
It can be seen that our method is capable of producing more precise 3D keypoints at the end of limbs, which further proves the effectiveness of our method.

\VspacePa
\section{Conclusion}
\VspacePb
\label{Conclusion}
This paper presents a Human Topology aware Network (HTNet), which takes full advantage of human structural priors for 3D HPE. 
To address the error accumulation, we design an Intra-Part Constraint (IPC) module that utilizes the topological constraints from intra-part parent nodes to reduce the errors of end joints.
Based on the IPC, the hierarchical mixer is further designed to learn joint-part-body representations via a channel-split progressive structure, allowing the HTNet to efficiently build hierarchical representations of the human topology.
Extensive experiments show that the proposed HTNet achieves state-of-the-art performance. 
We hope our work can inspire more researchers on skeleton-based tasks, \textit{e.g.}, action recognition, and 3D human mesh reconstruction.

{\small

\bibliographystyle{IEEEbib}
\bibliography{ref}
}

\end{document}